\documentclass[letterpaper, 10 pt, conference]{ieeeconf}  % Comment this line out if you need a4paper

\IEEEoverridecommandlockouts                              % This command is only needed if 
\usepackage{graphics} % for pdf, bitmapped graphics files
\usepackage{epsfig} % for postscript graphics files
\usepackage{times} % assumes new font selection scheme installed
\usepackage{amsmath} % assumes amsmath package installed
\usepackage{amssymb}  % assumes amsmath package installed
\usepackage{amsfonts}       % blackboard math symbols
\usepackage{amstext} % for \text macro
\usepackage{mathtools}
\usepackage{bbm}
\usepackage{float}

\usepackage{caption}
\usepackage{subcaption}
\usepackage{hyperref}       % hyperlinks

\usepackage[sort,nospace]{cite}
\usepackage[capitalize, nameinlink]{cleveref}

\usepackage{booktabs} 
\usepackage{makecell}
\usepackage{siunitx}
\usepackage{algorithm}
\usepackage[noend]{algorithmic}
\usepackage{threeparttable}

\newcommand{\CASE}[1]{\STATE \textbf{case} #1\textbf{:} \begin{ALC@g}}
\newcommand{\ENDCASE}{\end{ALC@g}}

\newcommand{\DEFAULT}{\STATE \textbf{default:} \begin{ALC@g}}
\newcommand{\ENDDEFAULT}{\end{ALC@g}}
\newcommand{\DEFAULTLINE}[1]{\STATE \textbf{default:} }
\usepackage[nolist]{acronym}
\usepackage[dvipsnames]{xcolor}

\usepackage{hyphenat} % For \hyp command
\usepackage{xspace}
\newcommand{\eg}{e.g.\@\xspace}
\newcommand{\ie}{i.e.\@\xspace}

\title{\LARGE \bf
Task-Adaptive Design of Modular Aerial Manipulators \\Under Airflow Exposure Constraints
}
\author{Mengguang Li and Heinz Koeppl% <-this % stops a space
\thanks{*This work has been funded by the LOEWE initiative (Hesse, Germany) within the emergenCITY center [LOEWE/1/12/519/03/05.001(0016)/72].}% <-this % stops a space
\thanks{The authors are with the Department of Electrical Engineering and Information Technology, Technische Universität Darmstadt, 64287 Darmstadt, Germany.
{\tt\small \{mengguang.li, heinz.koeppl\}@tu-darmstadt.de}}%
}
\begin{document}
\maketitle
\thispagestyle{empty}
\pagestyle{empty}
%%%%%%%%%%%%%%%%%%%%%%%%%%%%%%%%%%%%%%%%%%%%%%%%%%%%%%%%%%%%%%%%%%%%%%%%%%%%%%%%
\begin{abstract}
Aerial manipulation with multirotor platforms enables physical interaction in complex environments, but rotor-induced airflow remains a critical limitation for tasks involving airflow-sensitive targets or surroundings. This paper presents an optimization-based design framework for modular aerial manipulators that jointly considers task wrench feasibility, end-effector placement, and airflow exposure constraints. We first introduce a novel categorization of target-side airflow tolerance and formulate the corresponding exposure requirements as geometric constraints. To efficiently model rotor-induced airflow, we introduce a compact cone-sphere envelope that approximates the spreading structure of a quadrotor's airflow while preserving computational tractability for optimization. Building on this formulation, we propose a reconfiguration optimization that adapts a modular aerial manipulator to diverse task wrench requirements while enforcing both target-side airflow exposure and intra-platform airflow interference constraints. Unlike prior designs that assume a fixed end-effector location, the proposed framework optimizes the end-effector placement together with the platform configuration. Scalability experiments and ablation studies validate the effectiveness of the proposed framework.
\end{abstract}
\section{INTRODUCTION} \label{sec1}
Aerial manipulation aims to enable physical interaction in elevated or otherwise hard-to-reach environments. Multirotor platforms are particularly attractive for such tasks due to their maneuverability, hovering capability, and mechanical simplicity. Recent prototypes have demonstrated high-precision task execution in aerial manipulation scenarios \cite{10520237}. However, unlike ground-based manipulators, multirotor systems inevitably generate rotor-induced airflow, including inflow above the rotors and the wake flow beneath them, commonly referred to as downwash. This airflow can disturb the manipulated target, the surrounding environment, and even other parts of the aerial platform itself, thereby introducing additional design and control challenges \cite{KiranA-RSS-25}.

One important but underexplored aspect is the airflow tolerance on the target side. For tasks involving airflow-sensitive targets or environments, the platform must satisfy the required manipulation wrench while limiting rotor-induced airflow near the contact region and along the relevant flight path. Existing works have addressed this issue only partially. In \cite{7139851}, a non-interference sphere is introduced around the end-effector. While this provides a local protection region, it does not directly generalize to tasks requiring airflow constraints along the interaction trajectory. Long-reach aerial manipulators \cite{9088973} increase the standoff distance between the rotors and the target to reduce airflow interaction, but they do not explicitly optimize this distance or analyze how it affects the wrench feasibility of the overall aerial system. More recently, FlyingToolbox \cite{cao2025proximal} achieved centimeter-level docking accuracy by estimating disturbance forces using a feedforward neural network. Nevertheless, in such systems the end-effector may still operate within the near-field airflow of the platform. Overall, target-side airflow exposure requirements are rarely categorized or analyzed systematically, which makes it difficult to specify and enforce airflow constraints across different aerial manipulation tasks.

\begin{figure}[t]
    \centering
    \includegraphics[width=.9\linewidth]{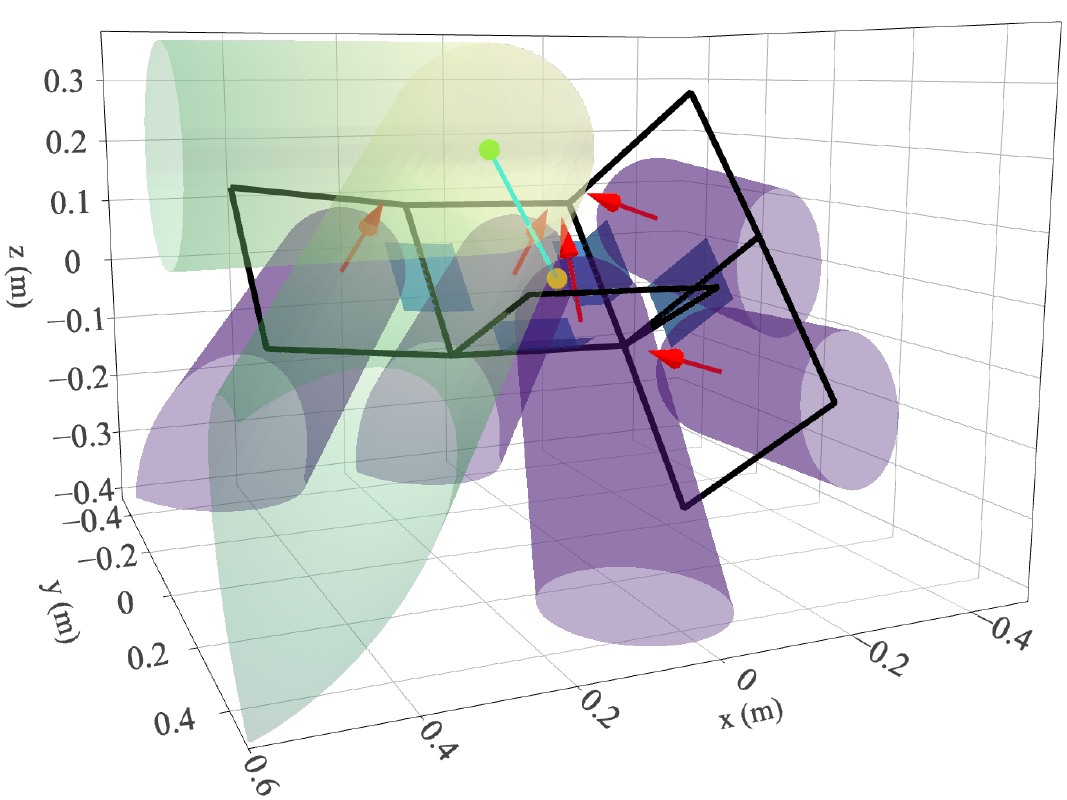}
    \caption{An optimal configuration of a five-module platform that satisfies the task in \cref{sec:airflow-cylinder} while respecting two target-side airflow requirements. Black squares represent modules, red arrows indicate each module’s $z$-axis, and blue rectangles denote the connected connectors. The green sphere marks the end-effector position, while the yellow sphere indicates the platform COM. The cyan line between them highlights the off-center end-effector placement. The purple and green-yellow volumes visualize the intra-platform and target-side airflow exposure envelopes, respectively.}
    \label{fig:fig1}
\end{figure}
Rotor-induced airflow is also critical on the platform side. Downwash interaction between rotors, modules, or neighboring aerial vehicles can reduce rotor efficiency, degrade tracking performance, and compromise stability \cite{KiranA-RSS-25}. For multirotor swarms, such interactions are often compensated at the control level, for example by using relative induced velocity models \cite{10804051} or learning-based residual modules \cite{11127714}. Although these methods can improve flight stability, avoiding strong airflow interference remains preferable when high motion precision and rotor efficiency are required. Several geometric airflow models have therefore been proposed for design and planning. An ellipsoidal non-collision volume is used in \cite{8202165}, but this representation can be overly conservative. An experimentally derived axisymmetric non-convex volume is introduced in \cite{7139851}. However, the associated iterative inner-loop optimization increases computational cost and limits scalability. A follow-up work \cite{11128060} approximates this volume using a discrete set of overlapping spheres. Other works use simplified cylindrical models \cite{9981140} or capsule-based airflow envelopes \cite{li2026downwash}, which provide computational convenience but do not fully capture the spreading behavior observed in recent studies \cite{10804051, KiranA-RSS-25}.

These observations motivate the need to incorporate rotor-induced airflow, both on the target side and within the aerial platform, into the optimal design of aerial manipulation systems. In addition to airflow constraints, the system must also satisfy the task-specific manipulation wrench requirement. Prior work on multirotor system design either does not explicitly consider wrench requirements \cite{8294249,10801469,9636086}, assumes that the desired wrench is applied directly at the center of mass (COM) \cite{10160555,02783649211025998,11146609,li2026downwash}, or treats an off-centered end-effector location as fixed a priori \cite{0278364919856694,drones10020129}. Consequently, a design that is feasible for one end-effector placement may fail to generate the same required wrench when the end-effector is mounted elsewhere. This indicates that end-effector placement, wrench feasibility, and airflow exposure should be treated as coupled design variables rather than independent post-design considerations.

In this work, we address these limitations by jointly considering wrench feasibility, end-effector placement, and airflow exposure constraints within a single nonlinear optimization framework. Building on the modular aerial platform introduced in \cite{li2026downwash}, we propose an optimal design method that reconfigures the aerial system according to task-specific wrench requirements while respecting both target-side and intra-platform airflow constraints. To model rotor-induced airflow, we introduce a cone-sphere envelope that provides a compact and realistic approximation of the spreading downwash region while retaining computational efficiency for optimization. Moreover, we relax the assumption that the end-effector placement is known a priori and instead optimize its location as part of the design. An illustrative example is shown in \cref{fig:fig1}.

The main contributions are summarized as follows:
\begin{itemize}
\item We propose a novel categorization of aerial manipulation based on target-side airflow tolerance, enabling systematic specification of airflow exposure constraints.
\item We introduce a cone-sphere airflow envelope for standard quadrotor modules, which captures the spreading structure of rotor-induced airflow while remaining suitable for efficient geometric constraint evaluation.
\item We formulate a joint optimization framework that simultaneously considers target-side airflow constraints, intra-platform airflow interference, end-effector placement, and task wrench feasibility for the design of modular aerial manipulation systems.
\end{itemize} 
\section{BACKGROUND AND SYSTEM MODEL}\label{sec:2}
\begin{figure}[t]
    \centering
    \includegraphics[width=.95\linewidth]{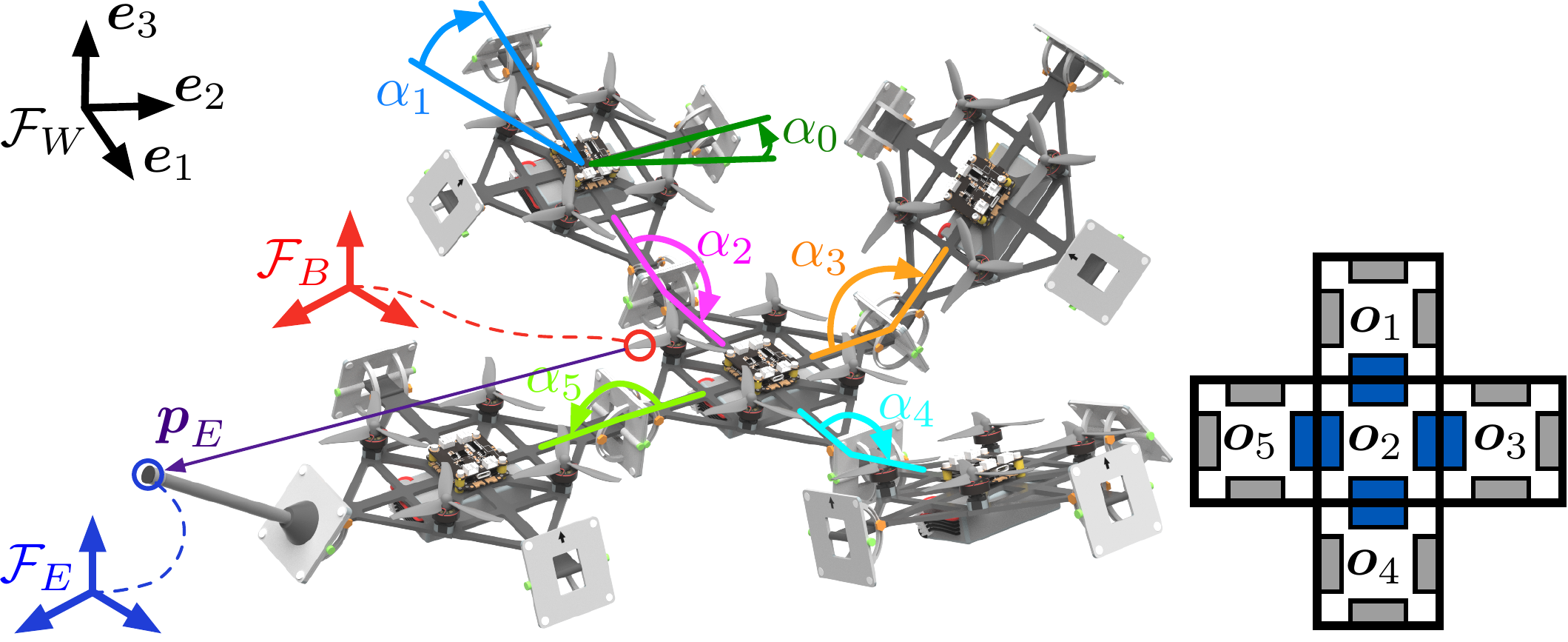}
    \caption{Left: An example five-module platform. The red circle marks the platform COM, and the blue circle marks the end-effector, here implemented as a rigid end-effector attached to one of the available connectors. Right: The corresponding representative graph, where gray rectangles indicate available connectors.}
    \label{fig:fig2_background}
\end{figure}

We use a standard quadrotor as the base module, which is a common baseline for modular multirotor aerial platforms. Each module is equipped with four connectors. Prior to assembly, each connector can be assigned a prescribed angle and then rigidly attached to a connector on another module. Following \cite{li2026downwash}, a platform is represented as an edge-labeled tree graph $\mathcal{G} = (\mathcal{V},\mathcal{E})$, where $\mathcal{V}$ denotes the set of modules and $\mathcal{E}$ the set of labeled edges, each encoding a connection with relative angle $\alpha$. An illustration is provided in \cref{fig:fig2_background}. We adopt the first step of their two-step pipeline to generate non-isomorphic configurations for $n$ modules. 

Let $\mathcal{F}_W$ denote the inertial frame and $\mathcal{F}_B$ the body frame attached to the platform COM. We consider a rigid end-effector and define the end-effector frame $\mathcal{F}_E$ such that it is co-aligned with $\mathcal{F}_B$. For a target end-effector wrench $[\boldsymbol{f}_E^{\top},\boldsymbol{\tau}_E^{\top}]^{\top}\in\mathbb{R}^6$, the corresponding wrench expressed about the COM in $\mathcal{F}_B$ is
\begin{equation*}
    \boldsymbol{f}_B = \boldsymbol{R}_E \boldsymbol{f}_E, \quad \boldsymbol{\tau}_B = \boldsymbol{R}_E \boldsymbol{\tau}_E + \boldsymbol{p}_{E} \times \boldsymbol{f}_B,
\end{equation*}
where $\boldsymbol{p}_E$ is the end-effector position relative to the COM. Since $\mathcal{F}_E$ and $\mathcal{F}_B$ are co-aligned, we set $\boldsymbol{R}_E=\boldsymbol{I}_{3\times 3}$. For a fixed end-effector wrench, the resulting wrench about the platform COM therefore depends on the end-effector placement $\boldsymbol{p}_E$. Given an optimal $\boldsymbol{p}_E$, the end-effector can be realized using one or more available connectors and rigid rods of appropriate lengths, as commonly used in off-center aerial manipulators \cite{0278364919856694}. One example realization is shown in \cref{fig:fig2_background}.

For a given graph $\mathcal{G}$ of an $n$-module platform, we initialize the root module at $\boldsymbol{p}_1=[0,0,0]^{\top}$ with orientation $\boldsymbol{R}_1 = \boldsymbol{R}_x(\alpha_0)\boldsymbol{R}_y(\alpha_1)$, where $\boldsymbol{R}_{x}(\cdot)$ and $\boldsymbol{R}_{y}(\cdot)$ denote elementary rotations about the $x$- and $y$-axes, respectively. Because the topology is a tree, standard traversal procedures (\eg, breadth-first search) can be used to compute each module pose from the relative connection angle between a parent and its child. Specifically, for a module $i$ with parent $h$,
\begin{equation*}
    \boldsymbol{R}_i=\boldsymbol{R}_{h}\bar{\boldsymbol{R}_{i}}, \quad
\boldsymbol{p}_i = \boldsymbol{p}_{h} + \boldsymbol{R}_{h}\boldsymbol{t}_i + \boldsymbol{R}_i\boldsymbol{t}_i,
\end{equation*}
where $\boldsymbol{t}_i$ is the vector from the $i$ module's COM to its connected connector to $h$, expressed in the module body frame. The relative rotation $\bar{\boldsymbol{R}}_i$ is selected according to the connection axis: $\bar{\boldsymbol{R}}_i=\boldsymbol{R}_x(\alpha_i)$ for a connection along the $y$-axis, and $\bar{\boldsymbol{R}}_i=\boldsymbol{R}_y(\alpha_i)$ for a connection along the $x$-axis. Let $m$ be the mass of each module and $m_E$ the end-effector mass, the total mass is $m_t = nm+m_E$. Denoting the end-effector position in $\mathcal{F}_B$ by $\boldsymbol{p}'_E \in \mathbb{R}^3 $, the COM of the full platform is 
\begin{equation*}
    \boldsymbol{p}_{\mathrm{COM}} = \frac{1}{m_t}\left(m_E\boldsymbol{p}'_E+m\sum_{i=1}^{n}\boldsymbol{p}_i\right).
\end{equation*}
After computing the pose $(\boldsymbol{p}_i,\boldsymbol{R}_i)$ of each module $i=1,\dots,n$, we shift the module positions and the end-effector position $\boldsymbol{p}'_E$ to the platform COM:
\begin{equation*}
    \boldsymbol{o}_i = \boldsymbol{p}_i-\boldsymbol{p}_{\mathrm{COM}}, \quad
    \boldsymbol{p}_E = \boldsymbol{p}'_E-\boldsymbol{p}_{\mathrm{COM}}.
\end{equation*}
For an X-layout quadrotor with arm length $l_{\mathrm{arm}}$, define the rotor position vectors in the module body frame as
\begin{align*}
\boldsymbol{\delta}_1 &= \frac{l_{\mathrm{arm}}}{\sqrt{2}}[1,-1,0]^\top,\quad
\boldsymbol{\delta}_2 = \frac{l_{\mathrm{arm}}}{\sqrt{2}}[-1,-1,0]^\top,\\
\boldsymbol{\delta}_3 &= \frac{l_{\mathrm{arm}}}{\sqrt{2}}[-1,1,0]^\top,\quad
\boldsymbol{\delta}_4 = \frac{l_{\mathrm{arm}}}{\sqrt{2}}[1,1,0]^\top.
\end{align*}
For module $i$ with position $\boldsymbol{o}_i$ and orientation $\boldsymbol{R}_i$ in the platform frame $\mathcal{F}_B$, the rotor positions in $\mathcal{F}_B$ are given by
\begin{equation*}
\boldsymbol{o}_{iq}=\boldsymbol{o}_i+\boldsymbol{R}_i\,\boldsymbol{\delta}_q,
\quad q\in\{1,2,3,4\}.
\end{equation*}
Given rotor angular speed $\omega_{iq}$, the generated thrust and torque are modeled as
\begin{equation*}
    f_{iq}=k_f\omega_{iq}^2, \quad \tau_{iq}=(-1)^q k_{\tau}\omega_{iq}^2,    
\end{equation*}
where $k_f$ and $k_{\tau}$ are the rotor coefficients. Let $\boldsymbol{e}_3=[0,0,1]^{\top}$ and $\boldsymbol{z}_i=\boldsymbol{R}_i\boldsymbol{e}_3$, the total force and torque about the COM are
\begin{equation*}
    \boldsymbol{F} = \sum_{i=1}^{n}\sum_{q=1}^{4} f_{iq}\boldsymbol{z}_i, \quad
    \boldsymbol{\tau} = \sum_{i=1}^{n}\sum_{q=1}^{4}
    \left(f_{iq}\boldsymbol{o}_{iq}\times \boldsymbol{z}_i
    + \tau_{iq}\boldsymbol{z}_i\right),
\end{equation*}
Stacking squared rotor speeds as the control input $\boldsymbol{u}_k=\left[\omega_{11}^2,\dots,\omega_{n4}^2\right]^{\top}$, with bounds $0 \leq \omega^2_{iq} \leq u_{\mathrm{max}}$, the achievable wrench $\boldsymbol{W}=[\boldsymbol{F}^{\top},\boldsymbol{\tau}^{\top}]^{\top}\in\mathbb{R}^6$ satisfies
\begin{equation*}
    \boldsymbol{W}=\boldsymbol{A}(\boldsymbol{\alpha},\boldsymbol{p}'_E)\boldsymbol{u}_k,\quad \boldsymbol{A}(\boldsymbol{\alpha},\boldsymbol{p}'_E)=\left[\boldsymbol{A}_{11},\ldots,\boldsymbol{A}_{n4}\right],
\end{equation*}
where each column $\boldsymbol{A}_{iq}\in\mathbb{R}^6$ is given by
\begin{equation*}
\boldsymbol{A}_{iq}=
\left[
\begin{array}{c}
k_f\,\boldsymbol{z}_i \\
k_f\,\boldsymbol{o}_{iq} \times \boldsymbol{z}_i + (-1)^q k_{\tau}\,\boldsymbol{z}_i
\end{array}
\right].
\end{equation*}
For a target wrench applied at the end-effector $\boldsymbol{b}_E = [\boldsymbol{f}_E^{\top},\boldsymbol{\tau}_E^{\top}]^{\top} $, the corresponding wrench about the platform COM is $\boldsymbol{b}_B(\boldsymbol{p}'_E)=[\boldsymbol{f}_B^{\top},\boldsymbol{\tau}_B^{\top}]^{\top}$. The wrench feasibility condition \cite{10160555,li2026downwash} requires that there exists a feasible control input $\boldsymbol{u}_k$ such that
\begin{equation*}
    \boldsymbol{A}(\boldsymbol{\alpha},\boldsymbol{p}'_E) \boldsymbol{u}_k = \boldsymbol{b}_B (\boldsymbol{p}'_E),
\end{equation*}
where $\boldsymbol{\alpha}, \boldsymbol{u}_k$, and $\boldsymbol{p}'_E$ are decision variables.
\section{METHODS} \label{sec:3}
Aerial manipulation tasks are commonly categorized by the degree of environment coupling into \textit{momentary coupling}, \textit{loose coupling}, and \textit{strong coupling} \cite{8059875,10520237}, which is consistent with established categorizations in ground manipulation. For multirotor platforms, an additional task aspect is the rotor-induced flow field \cite{KiranA-RSS-25}, which can also be critical for targets or their surroundings. Complementary to environment coupling, we propose a task categorization based on target-side airflow exposure requirements. Specifically, we model such requirements as a velocity-limited envelope in which the rotor-induced airflow speed must remain below a threshold. We consider three envelope types:
\begin{enumerate}
    \item \textbf{Local envelope}: a bounded region around the interaction area where only low-velocity airflow is allowed.
    \item \textbf{Directional envelope}: a region or union of regions aligned with one or more prescribed motion paths, \eg, approach/departure directions or given flight paths.
    \item \textbf{Exterior envelope}: a region defined outside a specified allowed volume, where the airflow speed must remain below the threshold throughout the exterior space.
\end{enumerate}

A \textit{local envelope} applies when targets are sensitive mainly during interaction, or when the platform is delivered to the workspace by other means, \eg, cable-suspended multirotor platforms \cite{8793592,11082553}. A \textit{directional envelope} applies when low airflow exposure must be maintained along motion paths before and after interaction. Many standard aerial manipulation tasks fall into this category \cite{10520237}. An \textit{exterior envelope} captures stricter target-side airflow requirements, \eg, operation near walls or corners, or tasks in which the entire aerial platform must remain within confined spaces \cite{9088973}. 

The proposed target-side airflow categorization can serve as a complement to environment coupling when benchmarking aerial manipulators. For example, picking up a lightweight object with a standard quadrotor may appear trivial when considering environment coupling alone, but can become challenging once target-side airflow exposure requirements are imposed. Note that, for manipulation tasks with no target-side airflow exposure requirement, only intra-platform airflow interference needs to be considered, if applicable.

\subsection{No target-side airflow requirement} \label{sec:airflow-none}
Based on the proposed target-side airflow categorization, we develop a design framework for a modular aerial manipulation platform that enforces the corresponding requirements. We first consider the baseline case where only intra-platform interference is relevant. 

\begin{figure}
    \centering
    \includegraphics[width=0.75\linewidth]{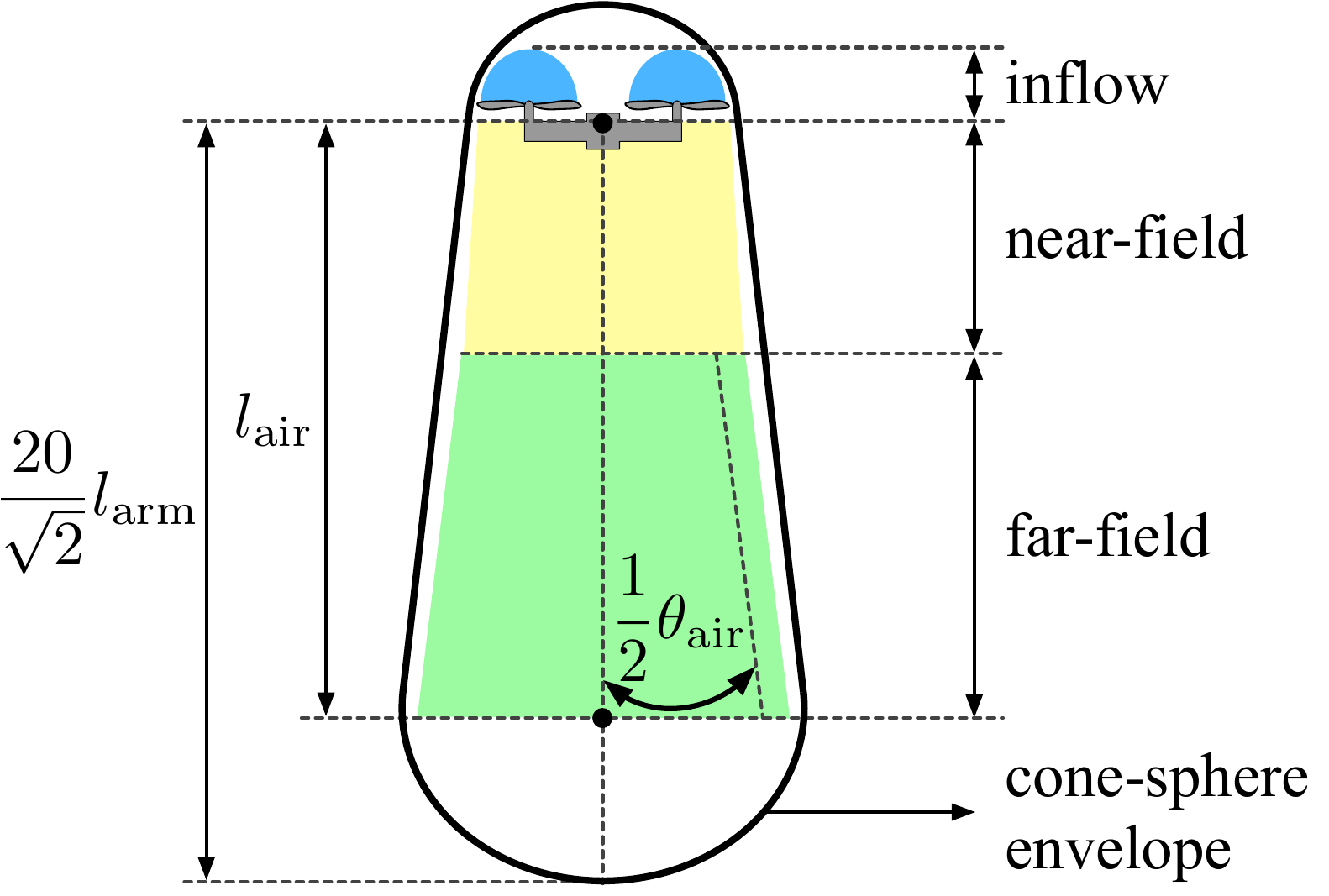}
    \caption{Illustration of the proposed cone-sphere envelope used to capture the rotor-induced airflow of a quadrotor. The envelope geometry is inspired by the empirical observations reported in \cite{KiranA-RSS-25} (Fig.~1). The envelope does not encompass the entire region with non-zero airflow velocity. Rather, it defines a critical region where the aerodynamic effects are considered non-negligible.}
    \label{fig:envelope_types}
\end{figure}

Most recently, a comprehensive experimental analysis of multi-quadrotor downwash was conducted using particle image velocimetry \cite{KiranA-RSS-25}. This study characterizes the quadrotor flow field and confirms that the near-field transitions approximately into a turbulent jet downstream in the far field \cite{10804051}. Motivated by these observations, we introduce a cone-sphere envelope to capture the dominant rotor-induced inflow, near-field, and far-field regions. The cone-sphere is constructed as a union of spheres swept along a line segment of length $l_{\mathrm{air}}$ with a radius that increases linearly along the segment. The radius growth $\Delta r_{\mathrm{air}}$ is chosen to match the turbulent jet spreading angle $\theta_{\mathrm{air}}$, yielding $\Delta r_{\mathrm{air}} = l_{\mathrm{air}}\tan(\theta_{\mathrm{air}}/2)$. With a sufficiently large $l_{\mathrm{air}}$, the airflow speed outside this envelope falls below a prescribed threshold. For demonstration in this work, we use the characteristic spreading angle $\theta_{\mathrm{air}}=\SI{10}{\degree}$ as in \cite{10804051}, and set the distance from module COM to the far end of the cone-sphere to $20(l_{\mathrm{arm}}/\sqrt{2})$. As reported in \cite{KiranA-RSS-25}, for distances larger than $19(l_{\mathrm{arm}}/\sqrt{2})$, the downwash effect becomes negligible, with the airflow speed outside the envelope dropping below \SI{40}{\percent} of the maximum induced velocity. An illustration is shown in \cref{fig:envelope_types}. The cone-sphere parameters can be adjusted to represent different levels of airflow enclosure, resulting in either a more conservative or less restrictive envelope model. 

Compared with the simplified capsule model in \cite{li2026downwash}, incorporating the cone-sphere model while maintaining comparable computational complexity is nontrivial. Enforcing non-intersection between each pair of cone-sphere envelopes naturally introduces two additional variables that parameterize the closest points on the corresponding centerlines. Similar to \cite{7139851}, this could be handled through an iterative inner-loop optimization within the overall design optimization. However, such an approach restricts scalability as the number of modules increases. Instead, we propose a smooth approximation of the minimum distance between two cone-sphere envelopes. This formulation avoids introducing two additional optimization variables for each module pair, while preserving smooth constraints suitable for efficient gradient-based optimization.

We begin by exploiting the geometric properties of the distance between two cone-sphere volumes. For module $i$ in the platform, the cone-sphere starts at its COM position $\boldsymbol{o}_i$ with radius $r_{\mathrm{air}}>0$ and extends along $\boldsymbol{d}_i=-l_{\mathrm{air}}\boldsymbol{z}_i$. A point on the segment $\boldsymbol{P}_i(\eta)$ and its sphere radius $r_i(\eta)$ can be formulated as 
\begin{equation*}
    \boldsymbol{P}_i(\eta) = \boldsymbol{o}_i+\eta \boldsymbol{d}_i,\quad
    r_i(\eta)=r_{i,0}+\Delta r_i\eta,
\end{equation*}
where $\eta \in [0,1]$, $r_{i,0}=r_{\mathrm{air}}$ and $\Delta r_i=\Delta r_{\mathrm{air}}$. Similarly, we define the cone-sphere line segment for module $j$ by $\boldsymbol{P}_j(\gamma)$ with $\gamma \in [0,1]$. The distance between the two cone-spheres can be written as 
\begin{equation*}
    d_{\mathrm{air}}(\eta,\gamma) =\| \boldsymbol{P}_i(\eta) - \boldsymbol{P}_j(\gamma)\| - (r_i(\eta) + r_j(\gamma)).
\end{equation*}
\begin{figure*}[t]
    \centering
    \includegraphics[width=.9\linewidth]{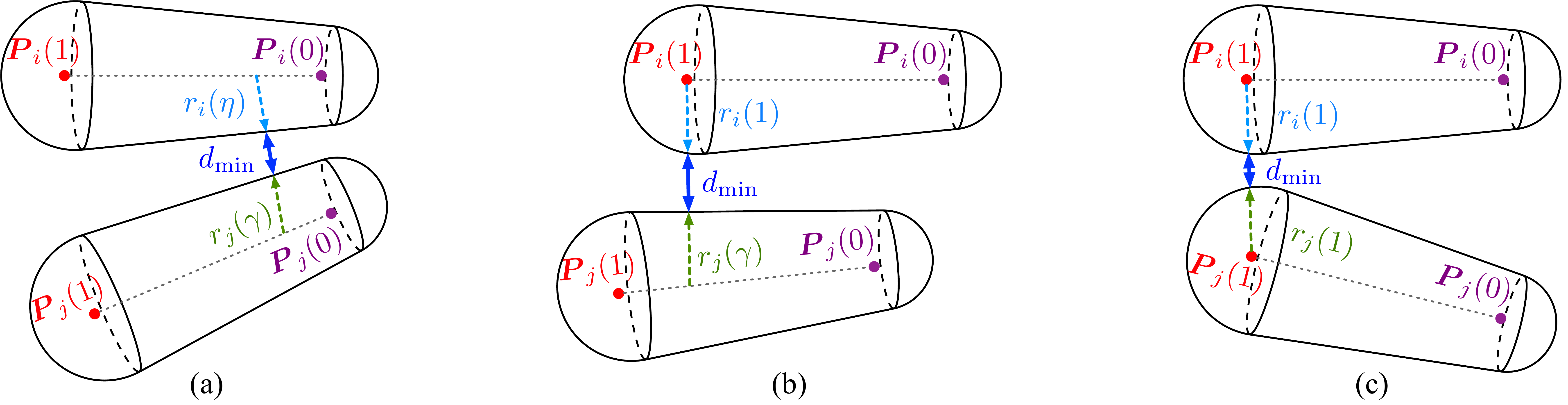}
    \caption{Illustration of all three cases for the minimum distance $d_{\mathrm{min}}$ between two cone-spheres, corresponding to the solution of \cref{eq: min_f_air}. (a) The minimizer of $f_{\mathrm{air}}(\eta,\gamma)$ is an interior stationary point, \ie, $0<\eta<1$ and $0<\gamma<1$, the minimum distance is attained between interior points of the two line segments. (b) The minimum distance is attained at one endpoint, shown here at $\boldsymbol{P}_i(1)$. (c) The minimizer is attained at two active bounds, shown here at the endpoints $\boldsymbol{P}_i(1)$ and $\boldsymbol{P}_j(1)$.}
    \label{fig:cone-spheres}
\end{figure*}

To ensure that the two volumes for modules $i$ and $j$ do not overlap, \ie $d_{\mathrm{min}} = \min \{d_{\mathrm{air}} \}\geq 0$, it is equivalent to requiring that the worst-case clearance is nonnegative, \ie,
\begin{align}
f_{\mathrm{air}}(\eta,\gamma) & \coloneqq \|\boldsymbol{P}_i(\eta) - \boldsymbol{P}_j(\gamma)\|^2-\big(r_i(\eta)+r_j(\gamma)\big)^2, \notag \\
\min_{(\eta,\gamma)\in[0,1]^2}\; & f_{\mathrm{air}}(\eta,\gamma) \geq 0.
\label{eq: min_f_air}
\end{align}
We observe that $f_{\mathrm{air}}(\eta,\gamma)$ is a bivariate quadratic polynomial and thus continuously differentiable. Since the feasible set $[0,1]^2$ is compact, a global minimizer exists. Moreover, any minimizer satisfies the Karush-Kuhn-Tucker (KKT) first-order necessary conditions for the bound-constrained problem \cite{nocedal2006numerical}. For the two-dimensional box constraints, the active set at a KKT point has three possibilities: 
\begin{enumerate}
    \item It is empty, in which case the minimizer is an interior stationary point with $0<\eta<1$ and $0<\gamma<1$;
    \item It contains one bound, in which case the minimizer lies on one of the four edges, \ie, $\eta\in\{0,1\}$ or $\gamma\in\{0,1\}$ with the other variable free;
    \item It contains two bounds, in which case the minimizer is one of the four corners $(\eta,\gamma)\in\{0,1\}^2$.
\end{enumerate}
For clarity, these cases are illustrated in \cref{fig:cone-spheres}. Accordingly, the global minimum of $f_{\mathrm{air}}$ on $[0,1]^2$ can be obtained by evaluating $f_{\mathrm{air}}$ on a finite set of analytically constructed candidates. Define $\boldsymbol{w}_0=\boldsymbol{o}_{i}-\boldsymbol{o}_{j}$, expanding $f_{\mathrm{air}}$ yields the quadratic form
\begin{equation}
f_{\mathrm{air}}(\eta,\gamma) = A\eta^2 + B\gamma^2 + C\eta\gamma + D\eta + E\gamma + F, \label{eq:f_air}
\end{equation}
with coefficients
\begin{align*}
A &= \|\boldsymbol{d}_i\|^2 - (\Delta r_i)^2, \quad
B = \|\boldsymbol{d}_j\|^2 - (\Delta r_j)^2,\\
C &= -2(\boldsymbol{d}_i^\top \boldsymbol{d}_j + \Delta r_i\,\Delta r_j),\\
D &= 2\,\boldsymbol{d}_i^\top \boldsymbol{w}_0 - 2(r_{i,0}+r_{j,0})\,\Delta r_i,\\
E &= -2\,\boldsymbol{d}_j^\top \boldsymbol{w}_0 - 2(r_{i,0}+r_{j,0})\,\Delta r_j,\\
F &= \|\boldsymbol{w}_0\|^2 - (r_{i,0}+r_{j,0})^2.
\end{align*}

The unconstrained stationary point is obtained by solving the linear system $\nabla f_{\mathrm{air}}(\eta,\gamma)=\boldsymbol{0}$. When the determinant $f_{\mathrm{det}}=4AB-C^2\neq 0$, this yields a unique stationary pair 
\begin{equation*}
    \eta_0 = \frac{(C  E - 2  B  D) f_{\mathrm{det}}}{f_{\mathrm{det}}^2 + \epsilon},\quad
    \gamma_0 = \frac{(CD - 2AE)f_{\mathrm{det}}}{f_{\mathrm{det}}^2 + \epsilon},
\end{equation*}
where $\epsilon>0$ is a small regularization term introduced for numerical stability when the stationary point is ill-conditioned, \ie, when $f_{\mathrm{det}} \approx 0$. We bound $(\eta_0,\gamma_0)$ to $[0,1]^2$ using the smooth approximations 
\begin{align*}
    \max \{a,b\} \approx \frac{1}{2}(a +b + \sqrt{(a-b)^2 + \epsilon}), \\
    \min \{a,b\} \approx \frac{1}{2}(a +b - \sqrt{(a-b)^2 + \epsilon}).
\end{align*}
The bounded pair $(\eta_0, \gamma_0)$ is the first evaluation candidate. 
\begin{figure}[t]
    \centering
    \includegraphics[width=.9\linewidth]{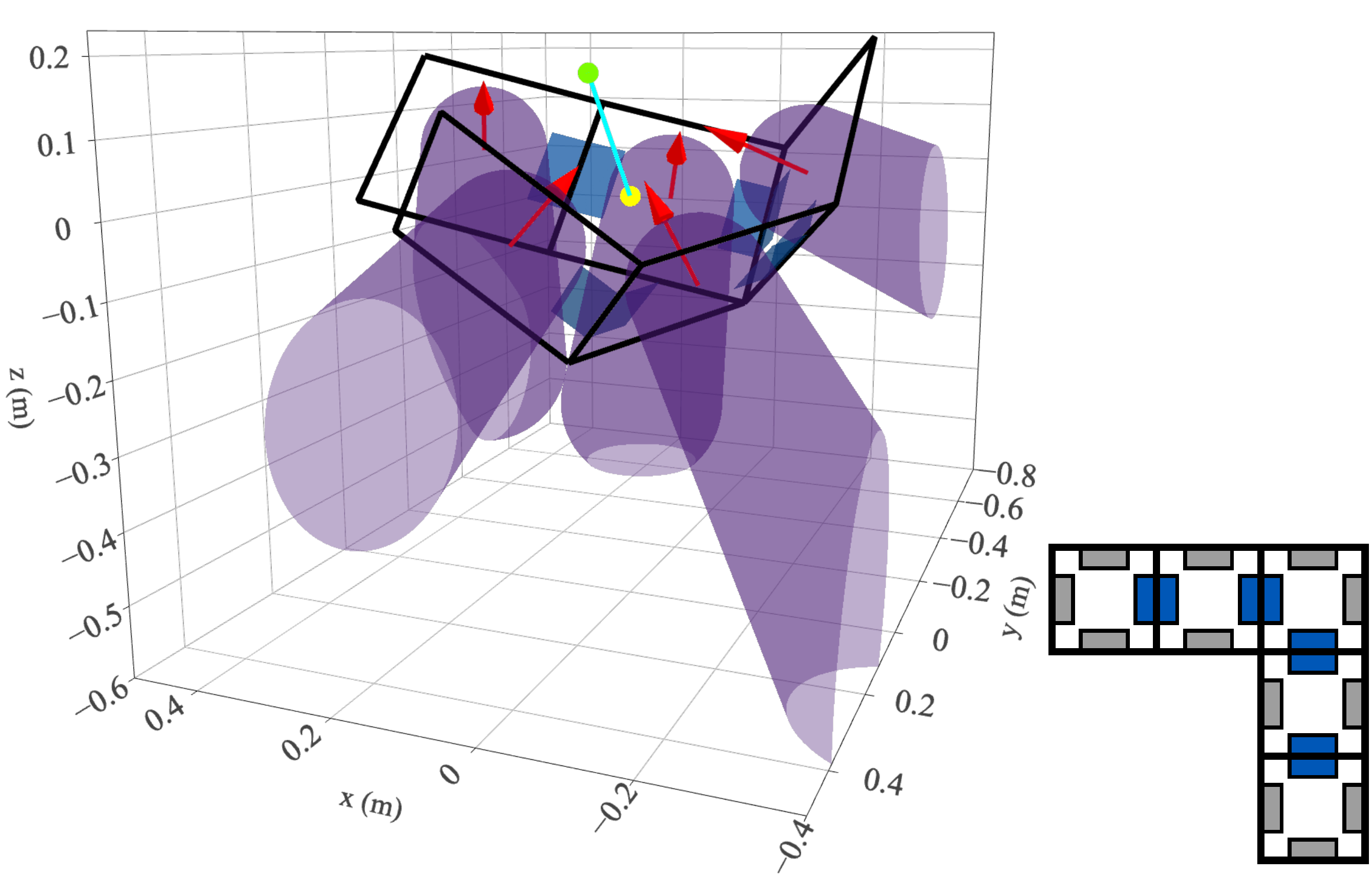}
    \caption{An optimal configuration of a five-module platform that satisfies wrench feasibility while respecting intra-platform airflow exposure constraints, with the corresponding representative graph shown on the right.}
    \label{fig:no_aero}
\end{figure}

If the global minimum lies on an edge, the restriction of $f_{\mathrm{air}}$ becomes a one-dimensional quadratic whose minimizer is given by the corresponding stationary point bounded in $[0,1]$. For the two edges $\eta_1 \in \{0,1\}$, we have 
\begin{equation*}
    \gamma_1=-\frac{(C\eta_1+E)B}{2B^2+\epsilon}.
\end{equation*}
For the other two edges $\gamma_2 \in \{0,1\} $, we have
\begin{equation*}
    \eta_2 = -\frac{(C \gamma_2 + D)A}{2A^2+\epsilon}.
\end{equation*}

We bound these four edge candidates in $[0,1]^2$ in the same manner. Finally, together with the four corners, we obtain nine candidates, producing a finite set $\mathcal{S}_{ij}\subset[0,1]^2$ for the module pair $i$ and $j$.
\begin{figure*}[t]
    \centering
    \begin{subfigure}[t]{.48\linewidth}
        \centering
        \includegraphics[width=.88\linewidth]{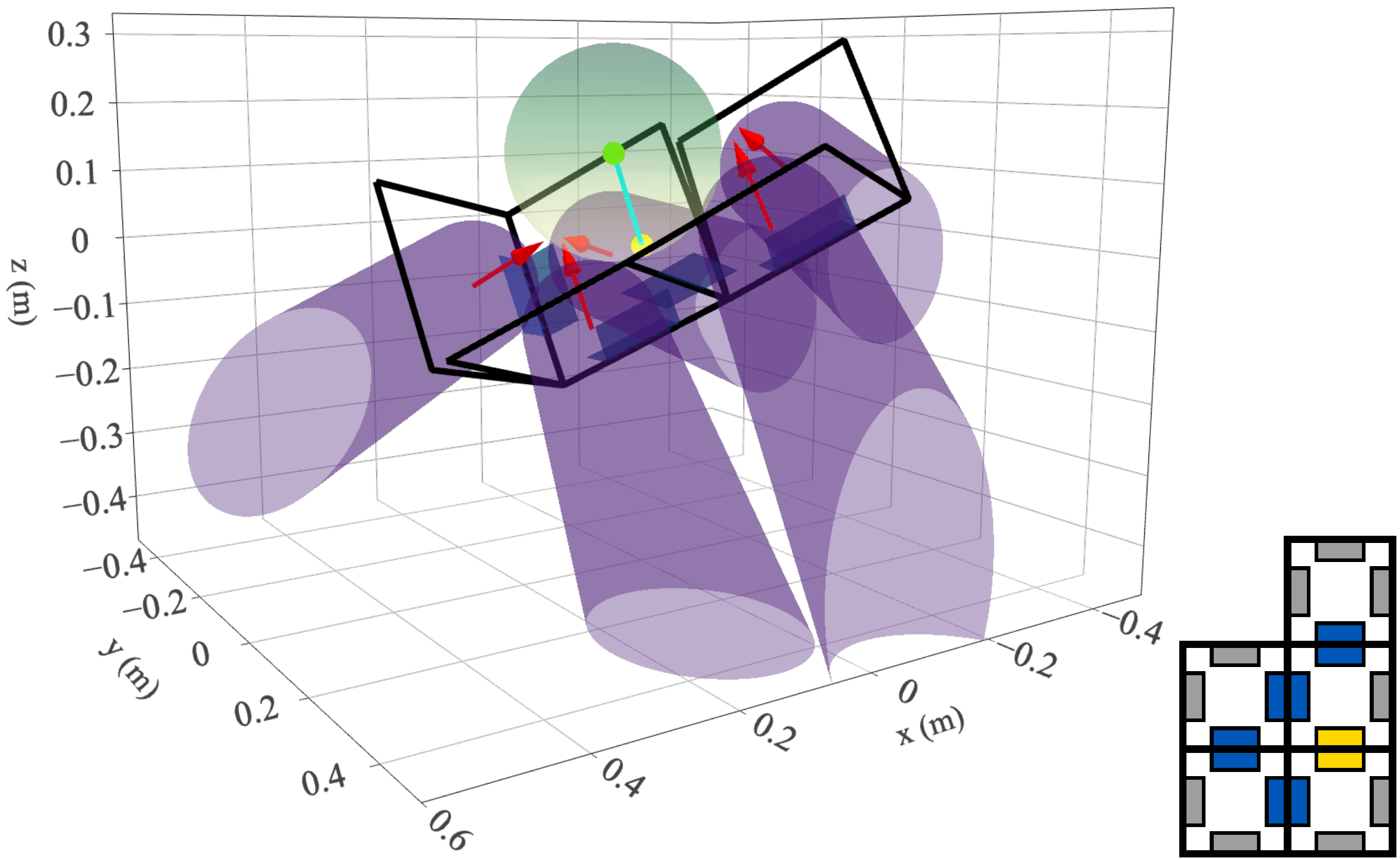}
        \subcaption{}
        \label{fig:aero_sphere}
    \end{subfigure}\hfill
    \begin{subfigure}[t]{.48\linewidth}
        \centering
        \includegraphics[width=.88\linewidth]{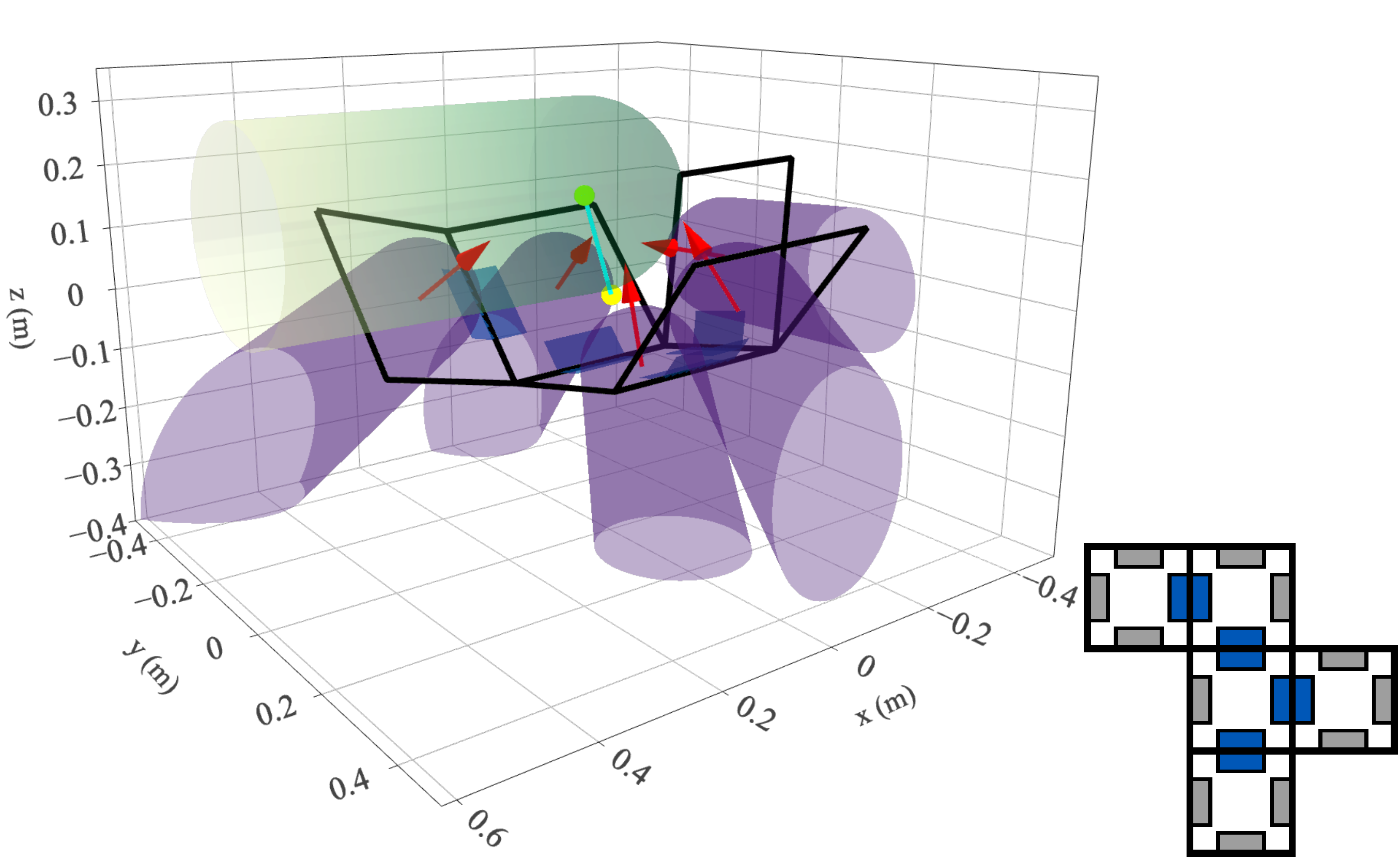}
        \subcaption{}
        \label{fig:aero_capsule}
    \end{subfigure}
    \begin{subfigure}[t]{.48\linewidth}
        \centering
        \includegraphics[width=.88\linewidth]{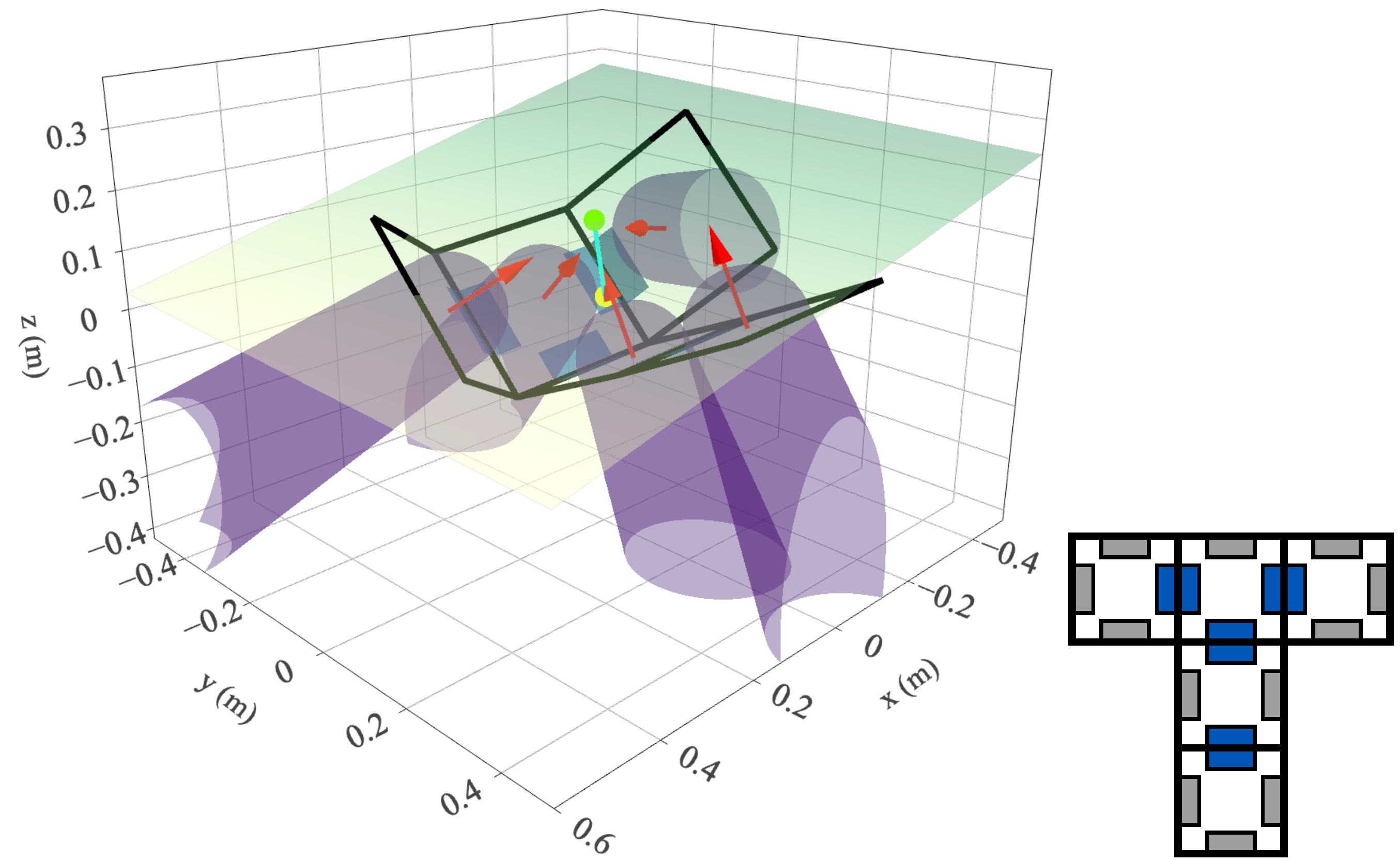}
        \subcaption{}
        \label{fig:aero_cone}
    \end{subfigure}\hfill
    \begin{subfigure}[t]{.48\linewidth}
        \centering
        \includegraphics[width=.88\linewidth]{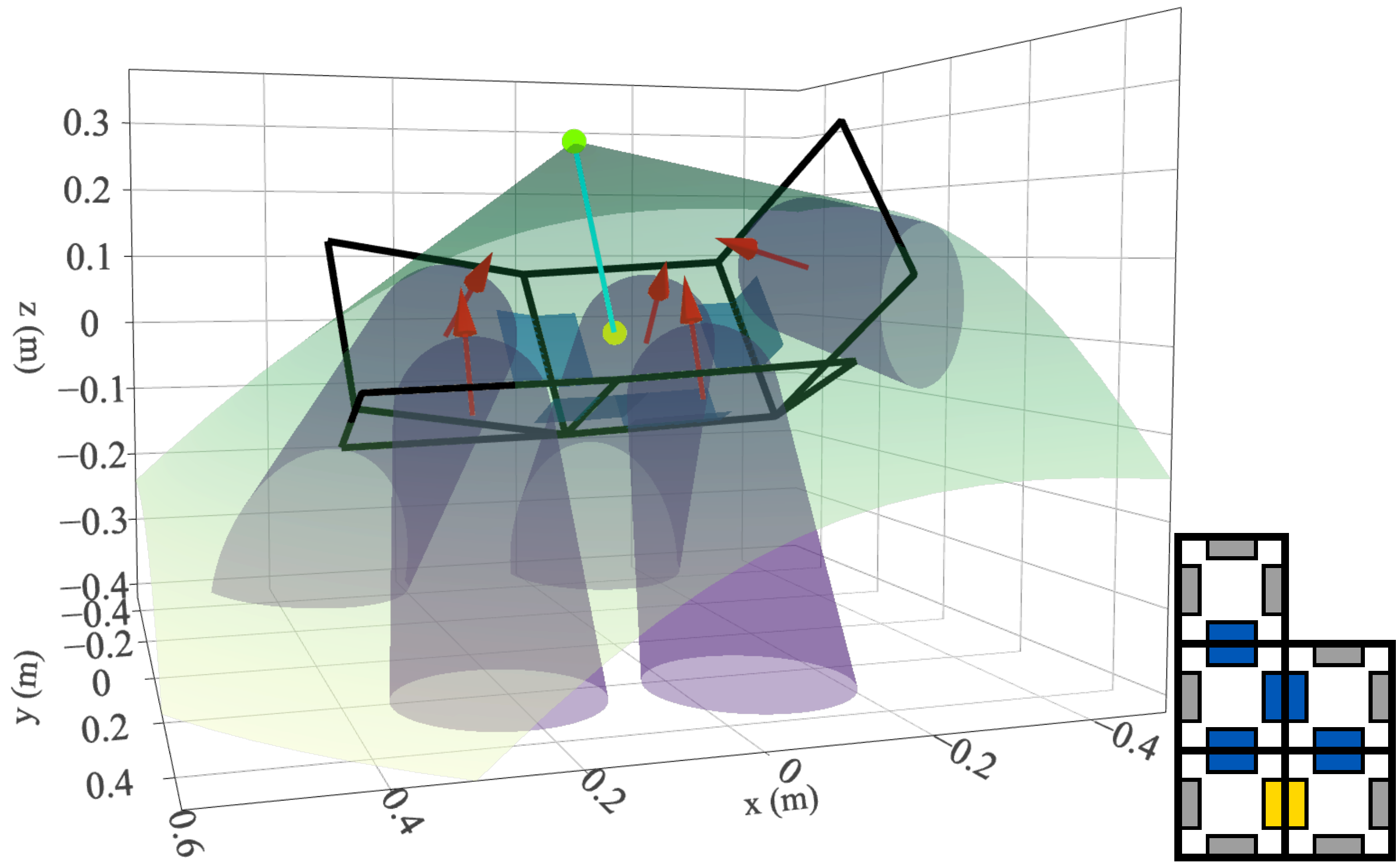}
        \subcaption{}
        \label{fig:aero_cone_2}
    \end{subfigure}
    \caption{Four examples using five-module platforms, with the corresponding configuration graphs shown on the right. Yellow rectangles indicate connectors that are not connected but are blocked for use. (a) An optimal configuration satisfying a target-side spherical local envelope constraint. (b) An optimal configuration satisfying a capsule directional envelope constraint with fixed segment direction $[1,0,0]^{\top}$. (c) An optimal configuration satisfying a half-space exterior envelope constraint with a parameterized axis $\boldsymbol{d}_{\mathrm{con}}$, where $\boldsymbol{p}_E$ is aligned with the plane normal. (d) An optimal configuration satisfying a conical exterior envelope constraint, where $\boldsymbol{p}_E$ is aligned with the cone axis.}
    \label{fig:aero_all_four}
\end{figure*}

To embed the clearance constraint into a gradient-based nonlinear program, we approximate the discrete minimum over a finite candidate set by a smooth log-sum-exp aggregation,
\begin{equation}
    \mathrm{LSE}_{\beta}\left(f,\mathcal{S}\right)
\coloneqq
-\frac{1}{\beta}\log\left(\sum_{\xi\in\mathcal{S}} \exp\big(-\beta f(\xi)\big)\right),
\label{eq:lse_def}
\end{equation}
where $\beta>0$ controls the approximation accuracy, larger $\beta$ yields a closer approximation to the true minimum while remaining smooth. Using \cref{eq:lse_def}, the airflow non-overlap constraints are enforced by
\begin{equation*}
\widetilde{\phi}_{ij}(\beta) =\mathrm{LSE}_{\beta}\left(f_{\mathrm{air}},\mathcal{S}_{ij}\right) \ge 0,\quad 1\le i<j\le n.
\end{equation*}
This construction avoids introducing additional decision variables $(\eta,\gamma)$ for each module pair and removes the need for inner-loop iterative distance computations, while providing a smooth constraint compatible with gradient-based methods.

The end-effector position $\boldsymbol{p}_E$ is allowed to vary in $\mathbb{R}^3$, subject to non-collision constraints with the platform. For each module $i$, we define a spherical exclusion region centered at its COM with radius $l_c$, yielding the constraint 
\begin{equation*}
    \| \boldsymbol{o}_i - \boldsymbol{p}_{E}\|^2 = \| \boldsymbol{p}_i - \boldsymbol{p}'_{E}\|^2 \geq l_c^2.
\end{equation*}

Let $\boldsymbol{\alpha} \coloneqq [\alpha_0,\dots,\alpha_n]^\top \in \mathbb{R}^{n+1}$. Together with the wrench feasibility conditions, inter-module airflow constraints, and end-effector non-collision constraints, we pose the modular reconfiguration design as the following nonlinear optimization problem: 
\begin{alignat}{3} \label{eq:opti}
\min_{\boldsymbol{u}_1,\dots,\boldsymbol{u}_K,\;\boldsymbol{\alpha},\; \boldsymbol{p}'_E} &\quad \sum_{k=1}^K \lambda_k\|\boldsymbol{u}_k\|^2_2 \\ \notag
\text{s.t. } &\quad \widetilde{\phi}_{ij}(\beta) \ge 0,\qquad \forall \; 1\le i<j\le n\\ \notag
&\quad \|\boldsymbol{p}_i-\boldsymbol{p}'_E \|^2 \geq l_c^2, \quad\forall \; i =1,\dots,n\\ \notag
&\quad \boldsymbol{A}(\boldsymbol{\alpha}, \boldsymbol{p}'_E) \boldsymbol{u}_k=\boldsymbol{b}_k (\boldsymbol{p}'_{E}), \quad \forall \; k =1,\dots, K\\ \notag
&\quad 0 \leq \boldsymbol{u}_k \leq u_{\mathrm{max}}, \quad \forall \; k = 1,\dots, K\\ \notag
&\quad -\pi/2 \leq \boldsymbol{\alpha} \leq \pi/2. \notag
\end{alignat}
Here, $\boldsymbol{u}_k$ and $\boldsymbol{\alpha}$ are bounded element-wise, $\boldsymbol{p}'_E \in \mathbb{R}^3$, and $\lambda_k > 0$ is a constant weight for task $k$.  

We use a simple task wrench set for illustration, two target wrenches applied at the end-effector, $\boldsymbol{W}_E = \{[0.5m_tg,0,0.5m_tg,0,0,0]^{\top},[0,0.5m_tg,0.5m_tg,0,0,0]^{\top}\}$, and an additional target wrench $[0,0,m_tg,0,0,0]^{\top}$ applied directly at the platform COM to enforce hovering capability. We solve \cref{eq:opti} with \texttt{Ipopt} 
\cite{wachter2006implementation,andersson2019casadi}. The optimal feasible solution using $5$ modules is shown in \cref{fig:no_aero}. Notably, the optimal end-effector position does not necessarily coincide with the platform COM. Because the objective minimizes overall control effort, an off-center placement can yield a lower cost for the same end-effector wrench than a placement at the COM.

\subsection{Local envelope} \label{sec:airflow-sphere}
To represent a \textit{local envelope}, a task-specific volume can be used, such as a sphere, box or other irregular shape. For our system, we choose a sphere of radius $ r_{\mathrm{sph}} > 0 $ that locally encloses the end-effector as the target-side airflow exposure constraint. This requires that the sphere centered at $\boldsymbol{p}_E$ does not intersect any cone-sphere envelope of the platform.

This case is a special instance of \cref{eq:f_air}, where one cone-sphere reduces to a single sphere. We take the cone-sphere of module $i$, keeping $\boldsymbol{P}_i(\eta) $ and $ r_i(\eta) $ unchanged, and set $\boldsymbol{P}_j(\gamma)=\boldsymbol{p}_E $ and $ r_j(\gamma)= r_{\mathrm{sph}}$. This yields a one-dimensional quadratic $f_{\mathrm{sph}}(\eta)$, and the candidate set reduces to three points: one stationary point obtained from $\nabla f_{\mathrm{sph}}(\eta) = 0$, and the two endpoints $\eta \in \{0,1\}$. Denote this set by $\mathcal{S}_{i,\mathrm{sph}}$. Using \cref{eq:lse_def}, we enforce the \emph{local envelope} exposure requirement by adding the following constraints to \cref{eq:opti}:
\begin{equation*}
    \widetilde{\phi}_{i,\mathrm{sph}}(\beta) = \mathrm{LSE}_{\beta}\left(f_{\mathrm{sph}},\mathcal{S}_{i,\mathrm{sph}}\right) \geq 0, \quad 1 \leq i \leq n. 
\end{equation*}
For the same wrench set in \cref{sec:airflow-none}, an optimal configuration example with $5$ modules is shown in \cref{fig:aero_sphere}. 

\subsection{Directional envelope} \label{sec:airflow-cylinder}
To adapt the modular platform to tasks with \textit{directional envelope} airflow constraints, we represent the prescribed motion path by a union of line segments. Sweeping the local sphere defined in \cref{sec:airflow-sphere} along each segment yields a capsule. Consequently, the overall \emph{directional envelope} is modeled as a union of capsules. 

Consider the cone-sphere envelope of module $i$ and one capsule in the \textit{directional envelope}. The corresponding non-intersection condition is another special case of $f_{\mathrm{air}}(\eta,\gamma)$, where the capsule has constant radius. Specifically, for a capsule starting at $\boldsymbol{p}_E$ with segment direction $\boldsymbol{d}_{\mathrm{cap}}$, we set
\begin{equation*}
   \boldsymbol{P}_j(\gamma)=\boldsymbol{p}_E + \gamma \boldsymbol{d}_{\mathrm{cap}}, \quad  r_j(\gamma)= r_{\mathrm{sph}}, 
\end{equation*}
which yields a bivariate quadratic $f_{\mathrm{cap}}(\eta,\gamma)$. The same KKT-based candidate enumeration as in \cref{sec:airflow-none} applies, resulting in nine candidates collected in $\mathcal{S}_{i,\mathrm{cap}}$. For each capsule, we add the following constraints using \cref{eq:lse_def} to the optimization problem \cref{eq:opti}:
\begin{equation*}
    \widetilde{\phi}_{i,\mathrm{cap}}(\beta) = \mathrm{LSE}_{\beta}\left(f_{\mathrm{cap}},\mathcal{S}_{i,\mathrm{cap}}\right)\geq 0, \quad 1 \leq i \leq n.
\end{equation*}
As an illustration, we set $\boldsymbol{d}_{\mathrm{cap}}=[1,0,0]^{\top}$ and solve the directional envelope problem for $5$ modules using the same target wrench set as in \cref{sec:airflow-none}. One optimal solution is shown in \cref{fig:aero_capsule}. Additional capsules can be incorporated by adding further segments, \eg, with $\boldsymbol{d}_{\mathrm{cap}}=[1,1,-1]^{\top}$, an example solution is shown in \cref{fig:fig1}. Note that the segment direction need not be specified a priori. For instance, it can be parameterized using the end-effector position, \eg, $\boldsymbol{d}_{\mathrm{cap}}=l_{\mathrm{cap}}\boldsymbol{p}_E/\|\boldsymbol{p}_E\|$, where $l_{\mathrm{cap}}$ is a scaling constant.

\subsection{Exterior envelope}
An \emph{exterior envelope} is used when low airflow exposure must be guaranteed throughout the surrounding workspace, while allowing rotor-induced flow only within a limited region. Similar to the previous two categories, the \emph{exterior envelope} can be instantiated according to the task and platform. As a representative realization, motivated by long-reach aerial manipulators \cite{9088973}, we constrain the platform's critical rotor-induced flow to remain inside an infinite cone with apex located at the end-effector. Equivalently, the \textit{exterior envelope} is defined as the space outside this cone. The cone half-angle controls the restrictiveness, a smaller half-angle reduces the allowed flow region near the target and provides stronger protection for the surroundings. However, tightening this constraint typically requires increasing the separation between the end-effector and the platform COM, which can reduce wrench feasibility. This trade-off between target-side airflow exposure and wrench feasibility is therefore central to the design of modular aerial manipulators, particularly for long-reach operation \cite{9088973}.

For a cone with apex at $\boldsymbol{p}_E$, axis given by the unit vector $\boldsymbol{d}_{\mathrm{con}}$, and half-angle $\theta_{\mathrm{con}} \in (0,\pi/2]$, we consider each sphere $\boldsymbol{P}_i(\eta)$ with radius $r_i(\eta)$ along the cone-sphere envelope of module $i$. Denote $\boldsymbol{q}_i(\eta) \coloneqq \boldsymbol{P}_i(\eta) - \boldsymbol{p}_E$, we have the orthogonal component of $\boldsymbol{q}_i(\eta)$ to the cone axis as 
\begin{equation*}
    \boldsymbol{\rho}_i(\eta) = \boldsymbol{q}_i(\eta)-\big(\boldsymbol{d}_{\mathrm{con}}^{\top}\boldsymbol{q}_i(\eta)\big)\boldsymbol{d}_{\mathrm{con}}.
\end{equation*} 
A sufficient condition for the sphere to lie inside the cone is that it does not extend behind the apex along the axis direction, and that it remains at least a distance $r_i(\eta)$ inside the conical surface. This can be enforced by
\begin{align}
    & \boldsymbol{d}_{\mathrm{con}}^{\top}\boldsymbol{q}_i(\eta) - r_i(\eta) \ge 0, \label{eq:cone_constraint_1} \\ 
& \boldsymbol{d}_{\mathrm{con}}^{\top}\boldsymbol{q}_i(\eta)\sin\theta_{\mathrm{con}}
    -\|\boldsymbol{\rho}_i(\eta) \| \cos\theta_{\mathrm{con}}
    -r_i(\eta) \ge 0. \label{eq:cone_constraint_2}
\end{align}
Since $\boldsymbol{P}_i(\eta)$ and $r_i(\eta)$ are affine in $\eta$, the left-hand side of \cref{eq:cone_constraint_1} is affine in $\eta$, and its minimum over $\eta\in[0,1]$ occurs at an endpoint $\eta\in\{0,1\}$. For $\theta_{\mathrm{con}} \in (0,\pi/2]$, we have $\cos\theta_{\mathrm{con}} \geq 0 $. Since the Euclidean norm is convex, the left-hand side of \cref{eq:cone_constraint_2} is therefore concave in $\eta$, hence its minimum also occurs at an endpoint. It is therefore sufficient to enforce \cref{eq:cone_constraint_1,eq:cone_constraint_2} by evaluating only the endpoints $\eta\in\{0,1\}$, and to include the resulting inequalities as constraints in \cref{eq:opti}. Similar to \cref{sec:airflow-cylinder}, the cone axis $\boldsymbol{d}_{\mathrm{con}}$ can be specified by the task as a fixed direction or parameterized based on the end-effector placement. In our examples, we set $\boldsymbol{d}_{\mathrm{con}} = - \boldsymbol{p}_E/\|\boldsymbol{p}_E\|$ and choose $\theta_{\mathrm{con}}=\SI{60}{\degree}$, as illustrated in \cref{fig:aero_cone_2}. By adjusting the cone half-angle $\theta_{\mathrm{con}}$, the restrictiveness can be tuned. In the limiting case $\theta_{\mathrm{con}}=\pi/2$, the cone boundary reduces to a plane, and an example solution is shown in \cref{fig:aero_cone}.
\begin{figure}[t]
    \centering
    \includegraphics[width=.95\linewidth]{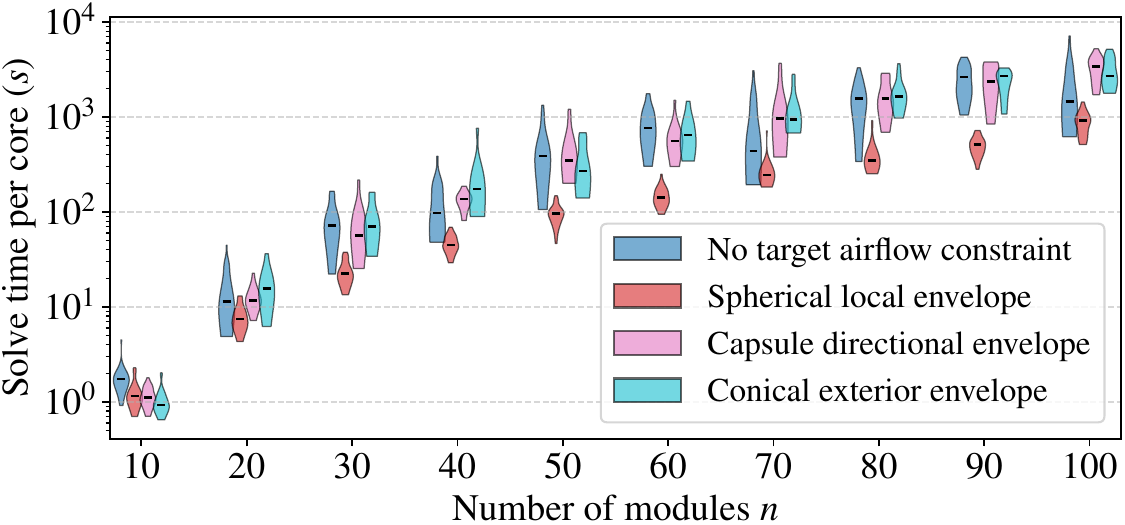}
    \caption{Scaling experiments under four constraint settings using $10$ target wrenches, reported every $10$ modules. For each $n$, the distributions are offset horizontally for visualization.} 
    \label{fig:exp1}
\end{figure}
\section{Experiments}\label{sec:4}
We first evaluate the scalability of the proposed framework. For each trial, we randomly generate $9$ target end-effector wrenches. The target forces along the $x$ and $y$ axes are sampled as $\pm \SI{20}{\percent}$ of the total weight $m_t g$, while the $z$-axis force is sampled from $[0.8m_t g,1.2m_t g]$. The target torques are sampled within $\pm \SI{20}{\percent}$ of the characteristic torque magnitude $l_{\mathrm{arm}}m_t g$. In addition, we include a gravity-compensation wrench applied at the platform COM in $\mathcal{F}_B$, $[0,0,m_t g,0,0,0]^{\top}$. For each module count from $10$ to $100$, we select $50$ non-isomorphic configurations as in \cite{li2026downwash} and evaluate the solver runtime on an AMD EPYC 7702P processor. Results are shown in \cref{fig:exp1}, indicating that the proposed method scales to large assemblies. We further observe that for a fixed number of modules the runtime remains in a similar range across different airflow constraint categories. 

\begin{figure}[t]
    \centering
    \includegraphics[width=1.\linewidth]{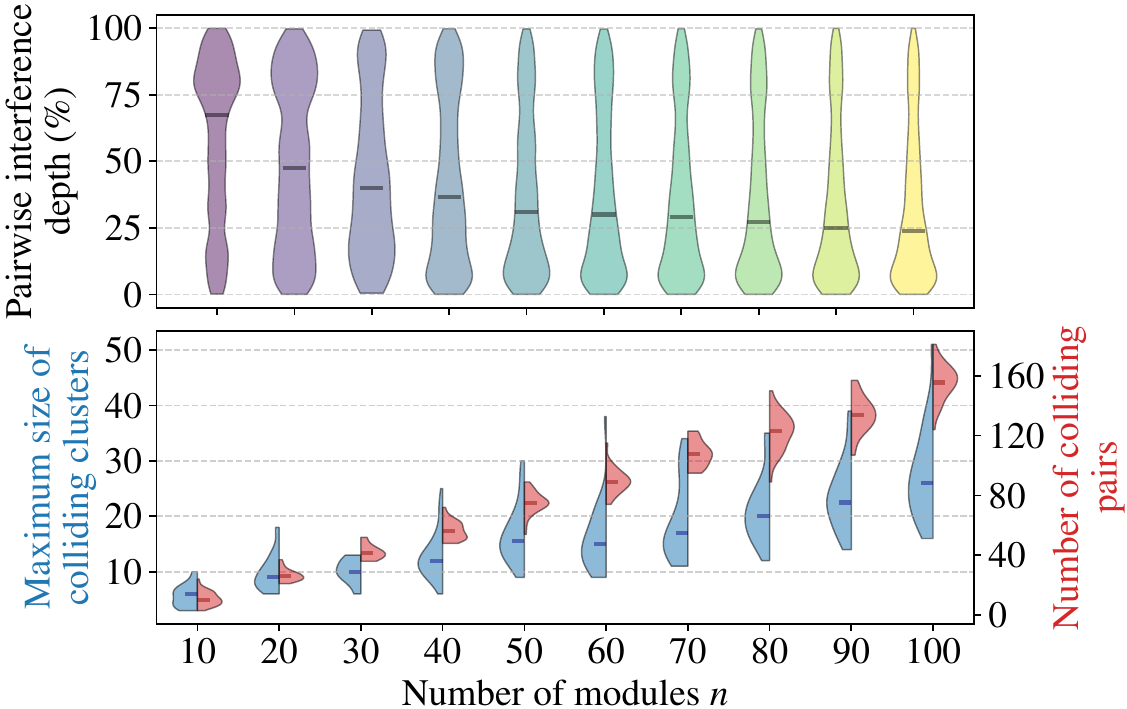}
    \caption{Ablation study of optimal configurations obtained without airflow constraints, using three metrics to quantify the severity of airflow interference.} 
    \label{fig:exp2}
\end{figure}

To further evaluate the effectiveness of the proposed airflow constraints, we conduct an ablation study in which the airflow non-overlap constraints are removed from \cref{eq:opti}. To avoid physical collisions between modules, we instead impose a minimum separation constraint $\|\boldsymbol{o}_i-\boldsymbol{o}_j\|^2 =\|\boldsymbol{p}_i-\boldsymbol{p}_j\|^2 \geq r_{\mathrm{m}}^2$. We report three metrics to quantify the severity of airflow interference. First, for each colliding module pair, we report the intersection depth as a percentage of the theoretical maximum possible overlap, $2(r_{\mathrm{air}}+\Delta r_{\mathrm{air}})$. Second, since airflow collisions may involve more than two modules, multiple cone-sphere envelopes can intersect and form a connected collision cluster. We therefore report the distribution of the maximum colliding cluster size over all trials. Third, we quantify how frequently airflow interference occurs by reporting the distribution of the number of colliding module pairs. The results are shown in \cref{fig:exp2}. We observe that the maximum interference level is frequently reached for different $n$, although the corresponding distributions vary across configurations. Moreover, large collision clusters, often involving approximately $n/2$ modules, occur in many trials. This indicates severe intra-platform aerodynamic interference when airflow constraints are omitted. In addition, the number of colliding pairs increases with $n$ and remains high across different configurations. Since airflow interaction between two quadrotors has been shown to affect the generated thrust and torque \cite{KiranA-RSS-25}, these results further motivate future experimental studies on airflow interactions involving multiple quadrotors.
\section{CONCLUSION} \label{sec:5}
In this work, we propose a target-side airflow exposure categorization for aerial manipulation. Building on this categorization, we develop an optimization-based design framework that jointly optimizes modular platform reconfiguration, end-effector placement, and both target-side and intra-platform airflow constraints. The resulting approach enables task-adaptive aerial manipulation while accounting for realistic rotor-induced airflow exposure requirements. In future work, we plan to build a physical aerial platform to evaluate the method in real-world experiments.

% \addtolength{\textheight}{0cm}   % This command serves to balance the column lengths
\bibliographystyle{IEEEtran}
\bibliography{references}
\end{document}